\documentclass[10pt,twocolumn,letterpaper]{article}

\usepackage{iccv}
\usepackage{times}
\usepackage{epsfig}
\usepackage{graphicx}
\usepackage{amsmath}
\usepackage{amssymb}
\usepackage{booktabs}
\usepackage{bbding}
\usepackage{multirow}

\usepackage{floatrow} 
\newfloatcommand{capbtabbox}{table}[][\FBwidth]
\usepackage{array}
\usepackage{enumitem}
\makeatletter
\newcommand{\thickhline}{%
    \noalign {\ifnum 0=`}\fi \hrule height 1pt
    \futurelet \reserved@a \@xhline
}
\newcolumntype{"}{@{\hskip\tabcolsep\vrule width 1pt\hskip\tabcolsep}}
\makeatother

\usepackage[pagebackref=true,breaklinks=true,letterpaper=true,colorlinks,bookmarks=false]{hyperref}

 \iccvfinalcopy 

\def\iccvPaperID{3637} 

\ificcvfinal\fi

\usepackage{hyperref}

\makeatletter
\newcommand{\printfnsymbol}[1]{%
  \textsuperscript{\@fnsymbol{#1}}%
}
\makeatother

\begin{document}

\title{Modular Adaptation for Cross-Domain Few-Shot Learning}

\author{Xiao Lin\thanks{Equal contribution} \quad Meng Ye\printfnsymbol{1} \quad Yunye Gong \quad Giedrius Buracas \quad Nikoletta Basiou \\ Ajay Divakaran  \quad Yi Yao \\ SRI International \\ Princeton, New Jersey, USA 08540}

\maketitle
\ificcvfinal\thispagestyle{empty}\fi

\newcommand{\paracompact}{\noindent\textbf}
\newcommand{\tb}{\textbf}
\newcommand{\red}{\textcolor{red}}

\begin{abstract}

Adapting pre-trained representations has become the go-to recipe for learning new downstream tasks with limited examples. While literature has demonstrated great successes via representation learning, in this work, we show that substantial performance improvement of downstream tasks can also be achieved by appropriate designs of the adaptation process. Specifically, we propose a modular adaptation method that selectively performs multiple state-of-the-art (SOTA) adaptation methods in sequence. 
As different downstream tasks may require different types of adaptation, our modular adaptation  enables the dynamic configuration of the most suitable modules based on the downstream task. 
Moreover, as an extension to existing cross-domain 5-way k-shot benchmarks (e.g., miniImageNet $\rightarrow$ CUB), we create a new high-way (\~{}100) k-shot benchmark with data from 10 different datasets. This benchmark provides a diverse set of domains and allows the use of stronger representations learned from ImageNet. Experimental results show that by customizing adaptation process towards downstream tasks, our modular adaptation pipeline (MAP) improves 3.1\% in 5-shot classification accuracy over baselines of finetuning and Prototypical Networks.

\end{abstract}
\section{Introduction}

\begin{figure*}[t]
    \centering

    \includegraphics[width=1.0\textwidth]{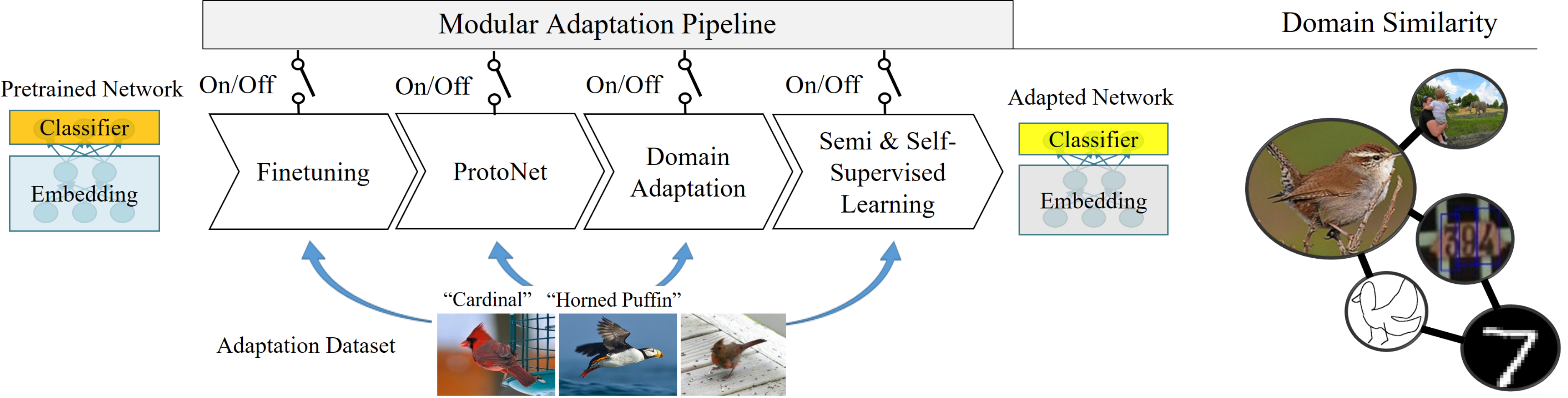}
    \caption{Block diagram of our proposed modular adaptation pipeline (MAP). MAP dynamically selects the most appropriate modules and their hyperparameters based on the characteristics of the given downstream task. MAP applies these selected modules in sequence to adapt the pre-trained models towards the target downstream task.}    
    \label{fig:teaser}
\end{figure*}

In few-shot learning (FSL), the task of building models using limited examples, adapting pre-trained representations has demonstrated successful applications in computer vision and natural language processing~\cite{fei2006one,vinyals2016matching,ravi2016optimization,hariharan2017low}. Finetuning pre-trained networks and learning classifiers on top of these embeddings leads to highly accurate classifiers even for data scarce conditions. An ideal pre-trained representation to start with is one that is learned in domains relevant to the target task or sufficiently diverse to enable effective transfer. However, in practice, relevant data is scarce and there is often a certain degree of domain shifts between the pretext and downstream tasks, where label ontology, viewpoint, image style, and/or input modality may differ. 
As such, cross-domain FSL~\cite{triantafillou2019meta,guo2020broader,tseng2020cross} recently brings renewed interest to this classical transfer learning problem focusing on the low data regime.

Existing studies show that depending on the characteristics of the underlying domain shifts, different downstream tasks may favor different adaptation methods, either finetuning-based~\cite{chen2019closer,guo2020broader} or metric learning-based approaches~\cite{tian2020rethinking,chen2020new}. 
For finetuning-based methods, the degree of finetuning required may also depend on the domain differences~\cite{li2020rethinking,cai2020cross,guo2019spottune} and the amount of training data available in the target domain~\cite{kolesnikov2019big}.
As a result, developing a one-size-fits-all cross-domain FSL approach has been challenging, if not entirely infeasible. To facilitate the appropriate design of cross-domain few-shot algorithms, a deeper understanding of domains, algorithms and their relationships is critical. 

In this work, we propose a modular approach for cross-domain few-shot adaptation that can be dynamically customized based on domain characteristics as an alternative to commonly used one-size-fits-all solutions. We refer to our method as the Modular Adaptation Pipeline (MAP, Figure~\ref{fig:teaser}). To build MAP, we first standardize few-shot adaptation methods into modules, where each module takes as the input a model and the adaptation datasets, and produces an adapted model as the output. 
We chain these modules into a consolidated pipeline, where multiple adaptation methods are applied in sequence. Given a downstream task, which adaptation modules should be applied as well as the hyperparameters of each selected module can be flexibly configured and automatically optimized via Bayesian optimization.

MAP can, thus, be considered as applying different loss terms according to the chosen methods sequentially to progressively adjust the input model towards the target downstream task. 
MAP enables flexible and customized integration of various SOTA techniques, whereas schemes based on joint optimization with an integrated loss function (e.g., weighted combination of multiple loss terms) may find it difficult to accommodate such diverse methods. Furthermore, while, in theory, our modular approach sacrifices global optimality due to step-wise optimization, schemes based on joint optimization, in practice, may also find it difficult to arrive at the global optima
simply because of the complexity of the loss landscape, which can be further complicated by the involvement of module hyper-parameters. As a result, our modular and sequential approach stands in contrast to methods based on joint optimization as an arguably better design choice for practical problems.  

As an exemplar realization of MAP, we integrate a collection of metric-based, finetuning, semi-supervised, and domain adaptation methods such as Prototypical Networks~\cite{snell2017prototypical,chen2020new}, batchnorm tuning~\cite{guo2020broader}, FixMatch~\cite{sohn2020fixmatch}, power transform, and semi-supervised embedding propagation~\cite{hu2020leveraging,ye2020hybrid}. This allows us to study their importance in different cross-domain few-shot adaptation tasks, providing empirical insights for a deeper understanding of adaptation algorithms, domain characteristics, and their relationships. 

Our major contributions include: 

1) We introduce a modularized approach for cross-domain few-shot adaptation to integrate a wide range of SOTA adaptation methods and apply them in sequence. In comparison to one-size-fits-all solutions, our method can dynamically select a sequence of the most suitable modules based on the characteristics of downstream tasks. In comparison to solutions based on joint optimization with combined losses, our method can work with a more diverse set of adaptation methods and achieve a balanced capability between finding the (sub)optimal solution and handling complex loss landscape. 

2) Our empirical analysis on the cross-domain performance of MAPs optimized for different domains shed insightful light on the similarities among downstream tasks, which varies depending on domain shifts and the availability of training data. 

3) We show that by customizing the adaptation process based on the target domain and availability of training examples, MAP produces on average a 3.1\% improvement in 5-shot classification accuracy over few-shot and finetuning baselines on a large-scale 100-way ImageNet $\rightarrow$ 10 datasets benchmark.

\section{Related work}
\subsection{Cross-domain few-shot learning}
 SOTA FSL approaches use a combination of representation learning for improving the pre-trained representation~\cite{tian2020rethinking,chen2020new,mangla2020charting,gidaris2019boosting,ye2020hybrid,chen2019self}, meta-learning for building more generalizable pre-trained models via episodic training~\cite{vinyals2016matching,finn2017model,snell2017prototypical}, and flexible adaptation such as prototype-based~\cite{ren2018meta,hu2020leveraging} and graph-based~\cite{satorras2018few,liu2018learning}. 
FSL benchmarks to date often split datasets into two disjoint sets of classes, one for pre-training and the other for adaptation. In practice, however, we often need to adapt SOTA pre-trained representations to learn classifiers using limited training examples on a different domain. Studies show that approaches designed for same-domain few-shot learning may not perform better than simple finetuning when there is a large domain gap between the pretext and downstream tasks~\cite{chen2019closer,guo2020broader}.
As such, recent works~\cite{triantafillou2019meta,guo2020broader,tseng2020cross} start to design more effective cross-domain FSL algorithms and more realistic cross-domain FSL benchmarks. 

Our modular adaptation approach focuses on cross-domain few-shot adaptation and advances existing work by providing a framework for integrating various adaptation methods. 
We show that cross-domain FSL approaches could benefit from customizing the adaptation algorithm towards the target domain through hyperparameter adaptation. 
We also introduce a larger-scale cross-domain FSL benchmark that adapts ImageNet pre-trained representations for downstream tasks with more classes (100-way vs. existing 5-way) to bridge the gap between existing small-scale cross-domain FSL benchmarks and the needed large-scale transfer learning.


\subsection{Adaptation methods in transfer learning}

In transfer learning, a variety of approaches for adaptating pre-trained classifiers to downstream tasks have been developed, such as fine-grained control of finetuning~\cite{li2018explicit,guo2019spottune} and network scaling~\cite{wang2017growing,kolesnikov2019big}. 
However, studies have shown that their performance depends heavily on hyperparameters, which may vary substantially according to the dataset size and the domain gap~\cite{li2020rethinking}. 

In contrast, our work shows that multiple adaptation methods and their hyperparameters can be configured under few-shot settings through cross validation and by transferring hyperparameters from similar tasks. This paves a distinctive path for integrating sophisticated adaptation methods in a more robust manner.

\subsection{Automatic transfer learning}

Inspired by the successes of automatic machine learning (AutoML), automatic transfer learning (i.e., automatically building models for new downstream tasks while leveraging models trained on pretext tasks) has gained increasing interests.
Relevant work includes automatically identifying relevant sources ~\cite{zamir2018taskonomy,Yan_2020_CVPR}, pre-trained model selection and empirical selection of hyperparameters for transfer learning~\cite{kolesnikov2019big,li2020rethinking}.

Our work focuses on the automated selection of adaptation methods along with their hyperparameters. The cross-domain performance of domain-specific MAPs can also be used as a post-hoc criterion for source selection. 


\section{Approach}

\begin{figure*}[t]
    \centering
    \includegraphics[width=1.00\textwidth]{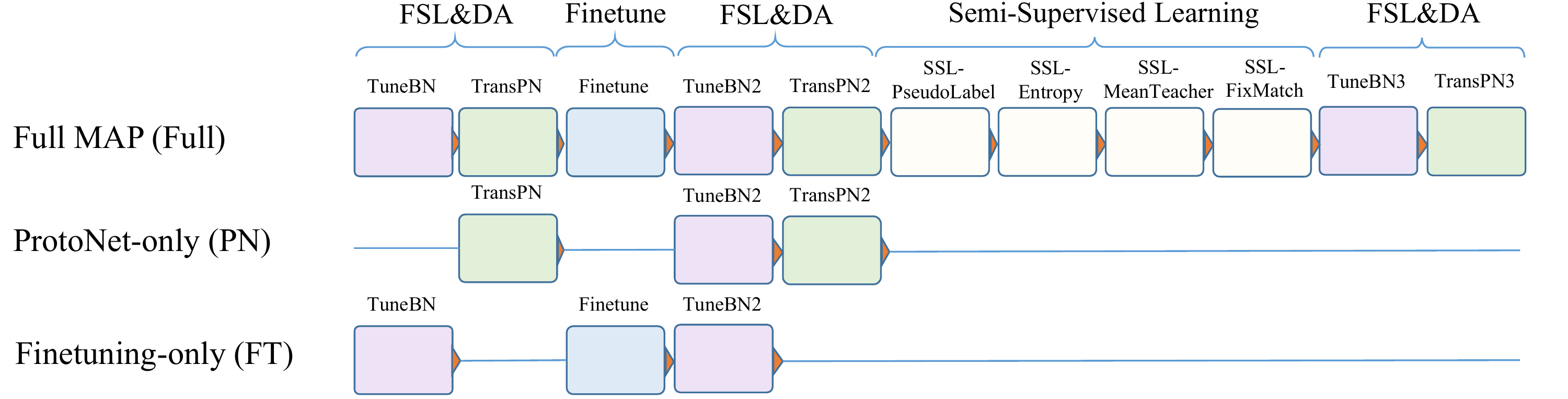}
    \caption{An exemplar realization of MAP based on seven adaptation methods of different types including FSL, domain adaptation (DA), finetuning and semi-supervised learning. These modules are chained sequentially with a fixed ordering and can be ``switched off'' using a skip connection. ProtoNet for FSL and finetuning for transfer learning can be implemented as special cases of MAP.}
    \label{fig:pipeline}
\end{figure*}


In this work, we focus on the problem of adapting a pre-trained model to perform image classification on a different dataset in a new domain (i.e., the target domain). Specifically, given 1) a pre-trained image classifier $F$ 
which takes in an image $I$ and outputs scores of classes $s=F(I)$, 2) an N-way K-shot labeled training set $D_l$, and 3) a set of unlabeled samples in the target domain $D_u$, the task is to build an updated image classifier $F'$ in the target domain to predict categories of testing samples. 

At the meta level, an adaptation approach can be abstracted into a module $F'=M(F,D_l,D_{u})$ or equivalently in operator form $F'=M(D_l,D_{u}) \circ F$. Multiple modules can be chained for a composite adaptation approach.
\begin{align*}
F' &= \underbrace{M_n (D_l,D_{u})\circ \cdots \circ M_1 (D_l,D_{u})}_{\text{n modules}} \circ F \\
&\triangleq M(D_l,D_u) \circ F
\end{align*}
Chaining multiple modules corresponds to adapting a pre-trained model using a collection of adaptation methods sequentially. In this work, we propose one such design, namely MAP, and study the benefits and challenges in constructing ``pipelined'' adaptation methods. 

We treat the configuration that switches on and off adaptation methods as additional hyperparameters to be learned during training. These configuration parameters are learned jointly with the hyperparameters of each selected module. While transfer learning and meta-learning have studied adapting model parameters to downstream tasks, adapting hyperparameters remains an under-explored area. 
However, hyperparameters should be adapted to the target domain, since studies indicate that the optimal effective learning rate in finetuning~\cite{kolesnikov2019big,li2020rethinking} and the effectiveness of different types of adaptation algorithms~\cite{triantafillou2019meta,chen2019closer,guo2020broader} both depend on domain similarity and the amount of training data. In this regard, MAP not only searches for the most suitable modules but also their hyperparameters for improved domain transfer. 

Combining multiple adaptation methods via learnable configuration leads to a large search space, where traditional gradient descend may find difficult to handle. 
To fully leverage the potential of MAP, we design a simple yet effective protocol to learn hyperparameters based on cross validation and transferring hyperparameters across tasks. 


In Section 3.1 and 3.2, we introduce the design space of our MAP including its building blocks and pipeline. We describe the proposed MAP search protocol in Section 3.3.

\subsection{Building blocks}

We implement seven modules based on adaptation methods with demonstrated successes in literature. The goal is to cover a diverse set of techniques such as FSL, domain adaptation and semi-supervised learning. A full list of options provided by each module is available in Appendix~\ref{sec:module_details}. 

\paracompact{Finetuning.} Finetune both the encoder (i.e., networks that map the input to an embedding feature space) and the classifier layers. The choice of optimizer (e.g., Adam or SGD), learning rate, momentum, weight decay, data augmentation, batch size, number of epochs and learning rate stepping schedule are set as hyperparameters. We also included the option of re-initializing the classification network with a fully connected layer. 

\paracompact{Transductive ProtoNet (TransPN).} Prototypical Networks with scaled cosine distance~\cite{chen2020new}, embedding power scaling~\cite{hu2020leveraging}, and CIPA~\cite{ye2020hybrid} prototype propagation for transductive learning. These algorithms have been shown to reach SOTA on FSL datasets. The cosine distance scaling factor, power scaling factor and the toggle of transductive learning are part of the hyperparameters. TransPN reduces the classification layer to simply comparing the embedding of a test sample against class prototypes based on scaled cosine similarity.

\paracompact{Finetuning batchnorm layer (TuneBN).} Updating batchnorm statistics with unlabeled data from the target domain helps model adaptaion in few-shot settings~\cite{nichol2018first,guo2020broader}. Network weights are frozen while batches of unlabeled data are passed through the network to update batchnorm statistics. TuneBN also serves the role of setting the momentum of batchnorm layers of the input network, which is an important factor in adaptation.

\paracompact{Semi-supervised learning with pseudo labels (SSL-PseudoLabel).} High-confidence predictions on unlabeled data are used as ``pseudo labels'' along with groundtruth labels during training~\cite{lee2013pseudo}. 

\paracompact{Semi-supervised learning with entropy minimization (SSL-Entropy).} Entropy on unlabeled examples is used as an additional loss term during training~\cite{grandvalet2005semi}.

\paracompact{Semi-supervised learning with student-teacher (SSL-MeanTeacher).} A semi-supervised learning approach that uses predictions from a running average network as the ``mean teacher'' to regularize the training of a student network~\cite{tarvainen2017mean}.

\paracompact{Semi-supervised learning with FixMatch (SSL-FixMatch).} Use consistency between strongly and weakly augmented inputs for semi-supervised learning~\cite{sohn2020fixmatch}.

\subsection{Pipeline design}

Figure~\ref{fig:pipeline} depicts an exemplar realization of our MAP. It consists of 11 modules. Finetuning is followed by semi-supervised learning with BatchNorm and ProtoNet modules in-between. Each module can be switched on or off. When switched off, a module is replaced with a skip connection, i.e., $F'=F$. 
While our empirical pipeline design does not cover all possible combinations of the building blocks, it supports a rich and compact set of configurations for selecting various methods. It covers standard baselines such as BatchNorm + ProtoNet and BatchNorm + finetuning. It also follows best practices such as finetuning the network before semi-supervised learning. 

The search space of hyperparameters consists of switches, $\{\text{on},\text{off}\}^n$, and the hyperparameters for each module, 126 hyperparameters in total. By keeping the number of modules in our pipeline fixed, standard Bayesian hyperparameter search techniques can be used to optimize MAP given a downstream task. 
Neural architecture search~\cite{zoph2016neural} and hyperparameter search transfer learning~\cite{perrone2018scalable} may enable further expansion of the search space and potentially result in better pipelines. In this work, we focus on studying the feasibility and impacts of MAP and leave the aforementioned further improvements to future work.

\begin{figure*}[t]
    \centering
    \includegraphics[width=1\textwidth]{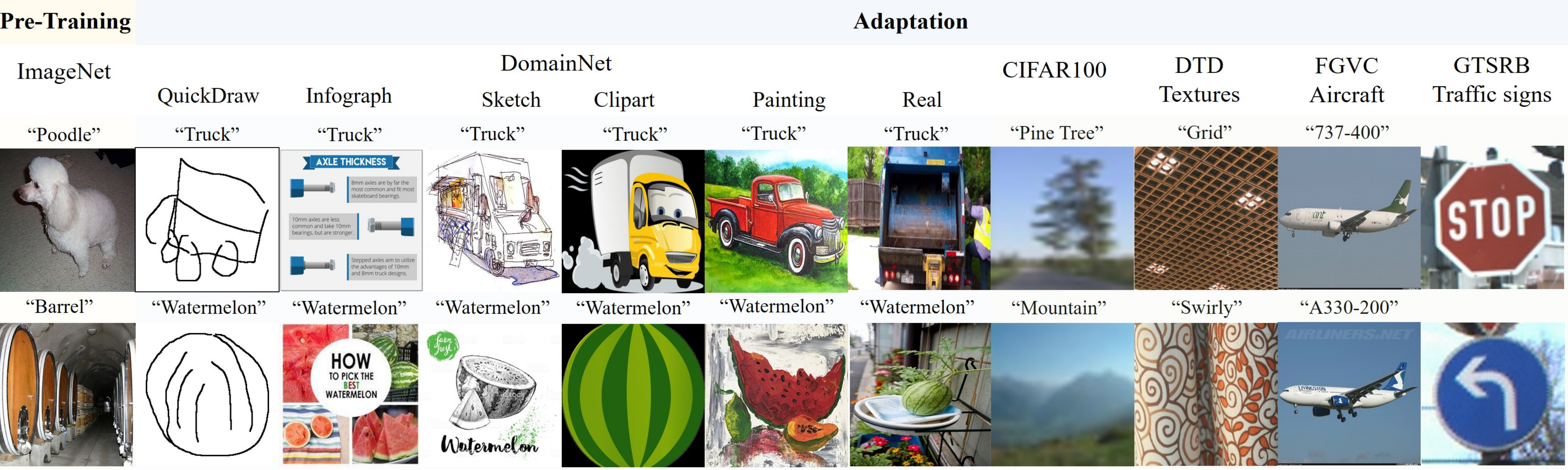}
    \caption{Example images in our ImageNet $\rightarrow$ 10 datasets cross-domain FSL benchmark. The datasets cover a wide variety of qualitatively different domains to assist the development of few-shot adaptation approaches.}
    \label{fig:dataset}
\end{figure*}

\subsection{Search protocol}

Given a downstream task $\{D_l,D_{u}\}$, the goal of domain dependent adaptation is to obtain an optimal set of MAP hyperparameters. The standard approach is to optimize $k$-fold cross validation performance: first construct multiple cross validation splits by randomly dividing $D_l$ into training splits $D^{train,i}_l$ and validation splits $D^{val,i}_l,i=1,\ldots,k$, and then perform adaptation on $\{D^{train,i}_l,D_u\}$ and optimize performance on $D^{val,i}_l$ using Bayesian hyperparameter optimization. Herein, the large search space presents a major challenge. For our exemplar MAP with 11 modules, there are $2^{11}=2,048$ possibilities in just deciding which modules should be turned on. From our experiments, we find that MAP optimization from scratch typically requires $>$300 iterations to converge, which often takes days on a single GPU. As reference, finetuning and ProtoNet, on the other hand, often converge within 30 iterations. 

To speed up MAP optimization, we propose to transfer MAP hyperparameters across domains. Intuitively, a MAP optimized for one adaptation problem may also perform well on similar problems with similar requirements. It, therefore, could serve as a better starting point of hyperparameter search than from scratch. A set of MAPs optimized for a diverse set of domains could readily cover a large variety of adaptation problems so that satisfactory performance could be reached by simply selecting the best performing MAP out of the set for a given new downstream task.

Specifically, we first collect an initial set of 20 to 50 high-performing MAPs on N-way K-shot adaptation tasks sampled from a diverse collection of datasets (excluding the target dataset), by running MAP hyperparameter search from scratch for 500 iterations. Building the initial collection of MAPs may take weeks across multiple GPUs. However, during adaptation, only MAPs in the collection are evaluated on the crossval splits $\{D^{train,i}_l,D_u\}$ and $D^{val,i}_l$ to select the most suitable MAP for the given downstream task. We show in experiments that transferring with a set of MAPs reaches similar accuracy as searching from scratch most of the times but converges more than 10$\times$ faster. 

Our search protocol is related to hyperparameter transfer learning~\cite{perrone2018scalable} and BiT-Hyperrule~\cite{kolesnikov2019big}, an empirical rule for determining finetuning hyperparameters. Different from~\cite{perrone2018scalable} which learns a transferable embedding space of hyperparameters, our approach simply finds hyperparameters that addresses a diverse set of tasks. MAP creates a more extended hyperparameter space than finetuning, which may open opportunities for deriving more sophisticated domain-dependent hyperparameter rules.

\section{Experiment}
\subsection{Experimental Setup}
\begin{table*}[t]
\centering
\footnotesize
\tabcolsep=0.11cm
\noindent

\begin{tabular}{ c | l |c c c c c c | c c c c | c }
	\toprule
	& Approach & QDraw & Infgrph & Sketch & Clipart & Pnting & Real & CIFAR & Textures & Aircraft & Signs & Overall  \\
\midrule
\multirow{3}{*}{2-shot} &	PN   & 21.37 &  8.35 & 21.59 & 35.60 & 36.89 & 66.68 & 32.33 & 40.62 & 11.85 & 27.30 & 30.26  \\
	                    &   FT   & 31.88 &  8.61 & 22.68 & 35.54 & 34.22 & 57.02 & 28.87 & 30.40 & 16.61 & 58.37 & 32.42  \\
                   &\textbf{MAP} & 30.10 &  9.91 & 27.38 & 39.25 & 39.66 & 69.25 & 35.63 & 41.19 & 14.01 & 57.42 & \textbf{36.38}  \\
\midrule
\multirow{3}{*}{5-shot} &	PN    & 30.51 & 12.72 & 31.82 & 48.09 & 45.72 & 72.39 & 42.85 & 52.00 & 18.28 & 39.21  & 39.36  \\
	                    &   FT    & 47.29 & 14.26 & 39.02 & 54.05 & 50.42 & 71.26 & 47.25 & 46.47 & 35.94 & 87.56  & 49.35  \\
                   &\textbf{MAP}  & 45.73 & 16.49 & 43.23 & 59.21 & 55.21 & 74.94 & 55.97 & 52.15 & 36.13 & 85.74  & \textbf{52.48}  \\
		\midrule
\multirow{3}{*}{10-shot} &  PN    & 35.23 & 16.75 & 40.85 & 54.99 & 51.44 & 75.50 & 49.25 & 56.04 & 22.86 & 44.74 & 44.77  \\
	                    &   FT    & 58.14 & 20.00 & 48.88 & 65.98 & 58.22 & 77.05 & 59.54 & 56.22 & 51.51 & 92.79 & 58.83  \\
                   &\textbf{MAP}  & 55.33 & 23.64 & 49.74 & 68.45 & 59.70 & 77.66 & 61.50 & 54.36 & 53.34 & 94.06 & \textbf{59.78}  \\
		\midrule
\multirow{3}{*}{20-shot} &  PN   & 39.62 & 21.36 & 44.60 & 60.77 & 56.48 & 77.91 & 53.94 & 56.30 & 26.79 & 53.49 & 49.13  \\
	                    &   FT   & 66.04 & 26.89 & 56.58 & 72.96 & 64.97 & 80.61 & 68.14 & 62.77 & 72.25 & 96.44 & 66.77  \\
                   &\textbf{MAP} & 62.89 & 30.28 & 55.65 & 75.23 & 65.25 & 79.93 & 68.42 & 61.12 & 75.05 & 98.37 & \textbf{67.22}  \\
		\bottomrule
\end{tabular}

\caption{Comparing performance of MAP against PN and FT baselines on 100-way K-shot ImageNet $\rightarrow$ 10 datasets cross-domain few-shot learning benchmark. MAP outperforms PN/FT baselines by \{3.96, 3.13, 1.04, 0.44\}\% at \{2, 5, 10, 20\}-shots, respectively. }

\label{table:transfer}
\end{table*}

\paracompact{Dataset.}
SOTA FSL has benefited from improving pre-trained representations on miniImageNet~\cite{vinyals2016matching,ravi2016optimization,triantafillou2019meta} through improved architectures~\cite{hariharan2017low,qiao2018few}, episodic meta-training~\cite{finn2017model,chen2020new}, knowledge distillation~\cite{tian2020rethinking} and self-supervised learning~\cite{gidaris2019boosting,mangla2020charting,ye2020hybrid}. Improvements on miniImageNet representations have been shown to be orthogonal to improvements in few-shot adaptation~\cite{hu2020leveraging,ye2020hybrid}, but are to certain degree disconnected from the representation learning community where the focus is on ImageNet~\cite{kolesnikov2019big,touvron2020training,radosavovic2020designing,chen2020simple} and beyond~\cite{dosovitskiy2020image}. 
To fully benefit from representation learning, we introduce a new large-scale 100-way 2- to 20-shot ImageNet $\rightarrow$ 10 datasets benchmark for cross-domain few-shot learning. 
These 10 adaptation datasets include DomainNet-\{real, clipart, sketch, quickdraw, infograph, painting\}~\cite{peng2019moment}, CIFAR-100~\cite{krizhevsky2009learning}, DTD-textures~\cite{cimpoi14describing}, FGVC-aircraft~\cite{maji13fine-grained} and GTSRB-traffic signs~\cite{Stallkamp2012}. Figure~\ref{fig:dataset} illustrates example images from the respective datasets. 
The choice of datasets is inspired by the VTAB ~\cite{zhai2019visual} and MetaDataset~\cite{triantafillou2019meta} benchmarks but with focus on fixed-way fixed-shot to study domain-dependent adaptation strategies.

All classes in the respective datasets are used except for DomainNet, where only the top 100 most frequent classes are used as there are insufficient testing examples for categories towards the tail of the class distribution. We randomly sample images to create K$\in$\{2,5,10,20\}-shot adaptation tasks with 20 test examples per class. The process is repeated over 5 random seeds to create 5 splits for each of the N-way K-shot problem. Accuracy is reported per-dataset per-shot, averaged over the 5 random splits. Following existing FSL conventions, unlabeled test examples are available for semi-supervised and/or transductive learning~\cite{ren2018meta,liu2018learning}. 
The splits are available at \url{https://github.com/frkl/modular-adaptation}.

We also benchmark our approach using existing cross-domain few-shot datasets including the VL3 challenge~\cite{guo2020broader} (5-way 5- to 20-shot miniImageNet $\rightarrow$ CropDisease~\cite{mohanty2016using}, EuroSAT~\cite{helber2019eurosat}, ISIC~\cite{codella2019skin} and ChestX~\cite{wang2017chestx}) and the four datasets from the LFT work~\cite{tseng2020cross} (5-way 1-\&5- shot miniImageNet $\rightarrow$ CUB~\cite{WahCUB_200_2011}, Cars~\cite{KrauseStarkDengFei-Fei_3DRR2013}, Places~\cite{zhou2014learning} and Plantae~\cite{van2018inaturalist}).

\paracompact{Backbone architecture.} We focus on adaptation from publicly available standard backbones for each benchmarking datasets. For VL3 and LFT, we used the baseline ResNet-10 pre-trained on miniImageNet provided by~\cite{cai2020cross}
\footnote{https://github.com/johncai117/Meta-Fine-Tuning} 
and ~\cite{tseng2020cross}
\footnote{https://github.com/hytseng0509/CrossDomainFewShot}
, respectively. For our ImageNet $\rightarrow$ 10 datasets settings, we used an EfficientNet-B0~\cite{tan2019efficientnet} pre-trained on ImageNet provided by the EfficientNet-pytorch model zoo\footnote{https://github.com/lukemelas/EfficientNet-PyTorch}. Input image resolutions are $224 \times 224$. 

\paracompact{Baselines.} We choose two baselines 1) finetuning (FT) where only TuneBN and finetuning modules are enabled and 2) prototypical networks (PN) where only TuneBN and Transductive ProtoNet modules are enabled as illustrated in Figure~\ref{fig:pipeline}. These two baselines are selected as their search space can sufficiently cover SOTA cross-domain few-shot adaptation techniques. 

\paracompact{Search strategy.} Hyperparameters are determined through 5-split cross validation on the adaptation set (each split with 50\% training and 50\% validation). As described in Section 3.3, to save computations of online adaptation
, we start from a set of pre-selected high-performing 1-, 2-, 5- and 10-shots pipelines on the 10 adaptation datasets\footnote{Using HyperOPT https://github.com/hyperopt/hyperopt}, total 40 pipelines for a reduced search space and, thus, decreased online computational complexity. Accordingly, we compare two hyperparameter selection strategies, one with the set of pre-selected pipelines (referred to as the \emph{transfer} setting), and the other with a set of best-performing pipelines optimized for lower shots on the adaptation dataset to approximate hyperparameter search from scratch (referred to as the \emph{from-scratch} setting). To verify the effectiveness of cross validation, we also design a third \emph{oracle} setting where hyperparameters are searched and evaluated directly on test data. For example, for 10-shot CIFAR-100 experiments, the \emph{transfer} setting selects pipelines from 1-, 2-, 5- and 10-shots from other datasets such as DomainNet-Clipart. The \emph{from-scratch} setting uses the pipelines optimized for 1-, 2-, and 5-shots on CIFAR-100. The \emph{oracle} setting uses the pipeline for 10-shot on CIFAR-100. 
 

\subsection{Experimental results}

\paracompact{Comparison against baselines.} Table 1 compares the performance of MAP vs. baselines (i.e., PN, FT) on the ImageNet $\rightarrow$ 10 datasets cross-domain task. Between PN and FT, we observe that PN, a FSL method, and FT, a transfer learning method, performs better on lower-shots (i.e., 2-shot) with similar domain and higher-shots (i.e., 5-, 10-, 20-shots) with disjoint domain, respectively, This agrees with the underlying designs of these methods and the general observations in literature. Thanks to its modularized design that supports online pipeline reconfiguration, MAP on average outperforms both PN and FT for 2-, 5-, 10-shots by 3.96\%, 3.13\% and 1.04\%, respectively. For 20-shot, MAP achieves comparable performance as FT. As the number of shots increases, the performance gain achieved by MAP degrades. This is expected since in general FT outperforms few-shot methods given sufficient labeled data.  
\begin{table*}
\centering
\scriptsize
\tabcolsep=0.11cm
\noindent

\centering
	\begin{tabular}{ l | c c | c c c c }
		\toprule
		Approach & Arch. & Trans. & CUB & Cars & Places & Plantae   \\
		\midrule
		RelationNet                    & ResNet-10 &             & 57.77 $\pm$ 0.69      & 37.33 $\pm$ 0.68      & 63.32 $\pm$ 0.76      & 44.00 $\pm$ 0.60 \\
RelationNet+LFT~\cite{tseng2020cross}  & ResNet-10 &             & 59.46 $\pm$ 0.71      & 39.91 $\pm$ 0.69      & 66.28 $\pm$ 0.72      & 45.08 $\pm$ 0.59 \\
		GNN                            & GNN       & \Checkmark  & 62.25 $\pm$ 0.65      & 44.28 $\pm$ 0.63      & 70.84 $\pm$ 0.65      & 52.53 $\pm$ 0.59 \\
		GNN+LFT~\cite{tseng2020cross}  & GNN       & \Checkmark  & 66.98 $\pm$ 0.68      & 44.90 $\pm$ 0.64      & 73.94 $\pm$ 0.67      & 53.85 $\pm$ 0.62 \\
		\midrule
		PN                             & ResNet-10 & \Checkmark  & 66.48 $\pm$ 1.08      & 51.68 $\pm$ 1.16      & 73.66 $\pm$ 1.04      & 58.98 $\pm$ 1.12 \\
		FT                             & ResNet-10 & \Checkmark  & 67.31 $\pm$ 1.03      & \bf{51.89 $\pm$ 1.14} & 71.68 $\pm$ 1.02      & \bf{60.26 $\pm$ 1.09} \\
		\textbf{MAP}                   & ResNet-10 & \Checkmark  & \bf{67.92 $\pm$ 1.10} & 51.64 $\pm$ 1.16      & \bf{75.94 $\pm$ 0.97} & 58.45 $\pm$ 1.15 \\
		\bottomrule
	\end{tabular}
%
%
\caption{Performance on 5-way 5-shot LFT benchmark.}

\label{table:LFT}
\end{table*}

\begin{table*}
\centering
\scriptsize
\tabcolsep=0.11cm
\noindent

\centering
	\begin{tabular}{ l |c c c c c c c c }
		\toprule
		Approach & \multicolumn{2}{c}{CropDisease} & \multicolumn{2}{c}{EuroSAT} & \multicolumn{2}{c}{ISIC} & \multicolumn{2}{c}{ChestX}  \\
		         & 5-shot & 20-shot & 5-shot & 20-shot & 5-shot & 20-shot & 5-shot & 20-shot   \\
		\midrule
		Transductive FT~\cite{guo2020broader}& 90.64 $\pm$ 0.54      & 95.91 $\pm$ 0.72      & 81.76 $\pm$ 0.48      & 87.97 $\pm$ 0.42      & 49.68 $\pm$ 0.36      & 61.09 $\pm$ 0.44      & 26.09 $\pm$ 0.96      & 31.01 $\pm$ 0.59      \\
		FT + Data aug.~\cite{cai2020cross}   & \bf{92.23 $\pm$ 0.46} & 95.95 $\pm$ 0.30      & 82.67 $\pm$ 0.50      & 87.84 $\pm$ 0.46      & \bf{51.76 $\pm$ 0.50} & 60.32 $\pm$ 0.59      & \bf{31.60 $\pm$ 0.41} & \bf{35.91 $\pm$ 0.42} \\
		\midrule
		PN                                   & 90.10 $\pm$ 1.11      & 93.95 $\pm$ 0.84      & 80.49 $\pm$ 1.27      & 86.08 $\pm$ 1.00      & 44.73 $\pm$ 1.31      & 56.86 $\pm$ 1.18      & 24.31 $\pm$ 0.83      & 29.48 $\pm$ 0.83      \\
		FT                                   & 85.52 $\pm$ 1.37      & \bf{96.07 $\pm$ 0.75} & 79.93 $\pm$ 1.31      & \bf{89.93 $\pm$ 1.29} & 46.86 $\pm$ 1.29      & \bf{62.86 $\pm$ 1.62} & 24.29 $\pm$ 0.85      & 30.60 $\pm$ 1.09      \\
        \bf{MAP}                             & 90.29 $\pm$ 1.56      & 95.22 $\pm$ 1.13      & \bf{82.76 $\pm$ 2.00} & 88.11 $\pm$ 1.78      & 47.85 $\pm$ 1.95      & 60.16 $\pm$ 2.70      & 24.79 $\pm$ 1.22      & 30.21 $\pm$ 1.78      \\
		\bottomrule
	\end{tabular}

\caption{Performance on 5-way 5- and 20-shot VL3 benchmark with ``single model finetuning''}

\label{table:VL3}
\end{table*}

In Table 1, DomainNet domains are arranged based on their realism with QDraw being the most abstract and Real the most realistic. Studying the performance across these domains, we notice that adaptation accuracy improves as the domain becomes more realistic. This is because the source domain is ImageNet. MAP achieves improved performance for all the domains except for QDraw, indicating that MAP's step-wise sequential optimization may be trapped into local optima at the earlier stage of the pipeline for scenarios with a large disparity between the source and target domains.  





We compare the performance of our MAP against existing approaches using the same backbone architectures on the LFT (Table~\ref{table:LFT}) and VL3 (Table~\ref{table:VL3}) cross-domain FSL benchmarks.\footnote{Additional implementation details are available in Appendix~\ref{sec:lft_vl3_details}.} We achieve competitive performance against SOTA on the small-scale 5-way datasets, especially for the LFT benchmark. Notably, on ISIC and ChestX in the VL3 benchmark, which are datasets of medical imagery, MAP performance is still relatively lower than the baselines. This is because our exemplar MAP does not include strong domain adaptation modules (the only domain adaptation comes from BatchNorm). By incorporating additional domain adaptation modules, we expect MAP performance to be further improved on these two datasets. 

\begin{table*}[hbt!]
\centering
\footnotesize
\tabcolsep=0.11cm
\noindent

\begin{tabular}{ l |c c c c | c c c c | c c c c }
	\toprule
	         & \multicolumn{4}{c}{From-scratch} & \multicolumn{4}{c}{Transfer} & \multicolumn{4}{c}{Oracle} \\
	Shots    &\bf{ MAP} & FT & PN & \bf{MAP vs FT/PN} & \bf{MAP} & FT & PN & \bf{MAP vs FT/PN} & \bf{MAP} & FT & PN & \bf{MAP vs FT/PN} \\
	\midrule
	2   & 36.15 & 31.36 & 31.81 & +4.34 &      35.93 & 31.66 & 31.75 & +4.18 &     36.34 & 30.74 & 32.30 & +4.04  \\
	5   & 47.70 & 44.14 & 40.77 & +3.56 &      49.14 & 46.05 & 40.21 & +3.09 &     49.00 & 46.28 & 41.06 & +2.72  \\
	10  & 55.84 & 54.14 & 46.61 & +1.70 &      55.75 & 54.71 & 45.79 & +1.04 &     55.92 & 55.07 & 46.61 & +0.85  \\
	20  & 61.30 & 61.22 & 50.34 & +0.08 &      61.54 & 61.34 & 50.12 & +0.20 &       -   &   -   &   -   &   -    \\
	\bottomrule
\end{tabular}


\caption{Comparing the performance of \emph{from-scratch} and \emph{transfer} hyperparameter selection strategies against the \emph{oracle} of 500 rounds of Bayesian hyperparameter optimization on 100-way K-shot ImageNet $\rightarrow$ DomainNet datasets. Both \emph{from-scratch} and \emph{transfer} can match the performance of \emph{oracle}. \emph{transfer} is more than 10x faster to run than \emph{from-scratch}.}

\label{table:summary}
\end{table*}

\paracompact{Comparison across search strategies.} Table 2 compares averaged performance of PN, FT, MAP across the six domains in DomainNet under the \emph{from-scratch}, \emph{transfer}, and \emph{oracle} settings for 2-, 5-, 10-, and 20-shots.\footnote{Accuracy on each domain is available in Appendix~\ref{sec:full_results}.} Our experiments under the \emph{oracle} setting show that the improvement margin between MAP and PN/FT on DomainNet is approximately $4.04\%$, $2.72\%$, and $0.85\%$, for 2-, 5-, and 10-shots, respectively. MAP under both \emph{from-scratch} and \emph{transfer} settings has achieved such improvement, suggesting that despite the large number of additional parameters introduced by dynamic pipeline configuration, MAP seems not yet suffer from over-fitting (i.e., similar improvement is achieved on the holdout fold of cross-validation). This allows for a larger search space by adding more modules into the MAP pipeline (e.g., self-supervised learning).   

\begin{figure}[t]
\centering
\includegraphics[width=0.75\textwidth]{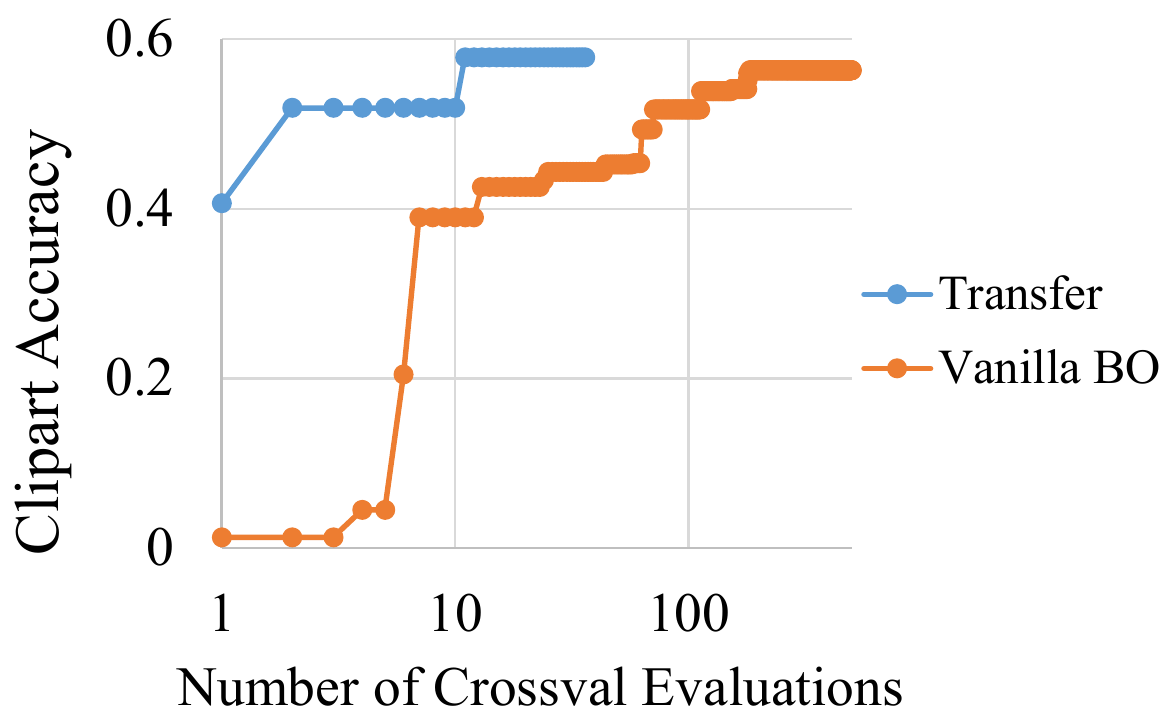}
\caption{MAP convergence rate on a DomainNet-Clipart 5-shot task. \emph{transfer} converges within 20 evaluations, whereas vanilla Bayesian hyperparameter optimization requires more than 200 evaluations to converge.}%
\label{fig:hp_convergence}
\end{figure}

Comparing between \emph{from-scratch} and \emph{transfer}, \emph{transfer} achieves similar performance to \emph{from-scratch}. The fact that the performance difference is marginal also verifies our conjecture that initializing from a set of pre-selected pipelines is a practical and effective alternative to searching the whole parameter space for reduced online computation. 
Figure~\ref{fig:hp_convergence} shows the convergence rate of \emph{transfer} against vanilla Bayesian hyperparameter search on 5-shot DomainNet-Clipart, where \emph{transfer} yields an approximately $20\times$ reduction in online computation.



\begin{figure}[t]
\centering
\includegraphics[width=1.0\textwidth]{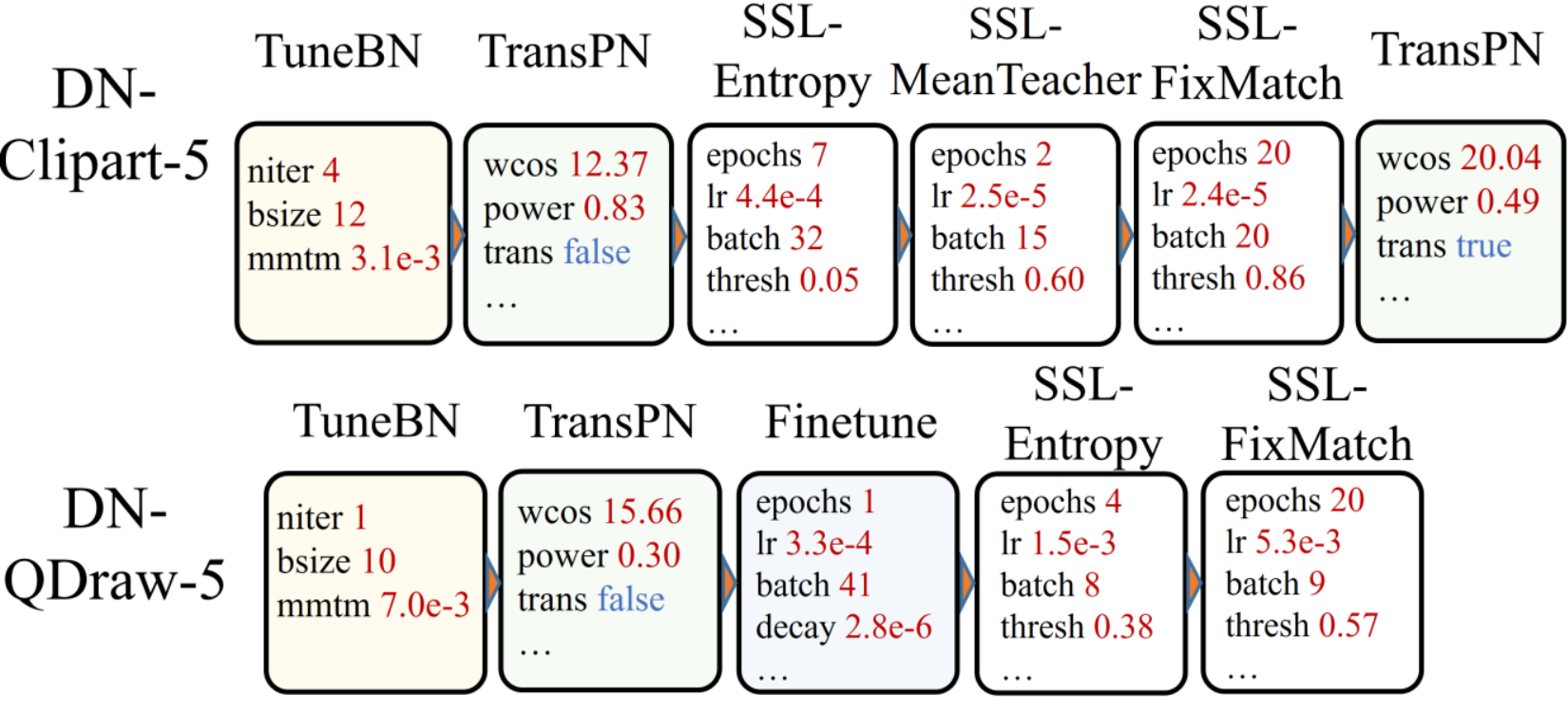}
\caption{MAPs optimized for 5-shot on DomainNet-Clipart (top) and DomainNet-QDraw (bottom).}%
\label{fig:pipelines_dn}
\end{figure}

\paracompact{Comparison across domains.} The domain-specific nature of MAPs allows for the study of which domains favors which adaptation methods. 
Table~\ref{table:pipeline_generalization} shows the cross-domain performance of six MAPs optimized for 5-shot on Domain-QDraw, -Infgrph, -Sketch, -Clipart, -Pnting, and -Real. In general, the best performance is achieved using the MAP optimized for the same or similar domains. Figure~\ref{fig:pipelines_dn} visualizes the MAPs optimized for DomainNet-Clipart-5 and DomainNet-QDraw-5. Both pipelines start from a ProtoNet initialization and perform semi-supervised finetuning. However, DN-Clipart-5 seems to emphasize FSL with TransPN as the output classifier, whereas DN-QDraw-5 relies more on finetuning with larger learning rates. This again verifies our conjecture that different domain characteristics favor different adaptation techniques and the need of adaptive configuration of these methods. 


\begin{table}[t]
\centering
\footnotesize
\tabcolsep=0.11cm
\begin{tabular}{ l |c c c c c c }
	\toprule
	MAP  & QDraw & Infgrph & Sketch & Clipart &  Pnting & Real \\
	\midrule
	DN-QDraw-5       & \bf{47.75} &      9.05  &     26.55  &     39.00  &     25.95  &     25.95      \\
	DN-Infgrph-5     &     38.15  &     14.80  &     38.15  &     48.80  &     46.40  &     70.45      \\
	DN-Sketch-5      &     42.65  &     17.20  & \bf{41.75} &     54.65  &     50.55  &     71.95      \\
	DN-Clipart-5     &     38.65  & \bf{17.50} &     41.45  &     56.75  & \bf{53.45} &     73.65      \\
	DN-Pnting-5      &     37.75  &     14.45  &     38.40  & \bf{56.80} &     52.40  & \bf{75.20}      \\
	DN-Real-5        &     33.95  &     16.05  &     39.05  &     54.10  &     49.70  &     73.15      \\
	\bottomrule
\end{tabular}
\caption{Cross-domain performance of six MAPs for 100-way 5-shot ImageNet $\rightarrow$ DomainNet tasks. 
}


\label{table:pipeline_generalization}

\end{table}

The cross-domain performance of these MAPs with high domain specificity by design can be used to measure domain similarity. In this regard, we collected MAPs optimized on ImageNet $\rightarrow$ 10 datasets for 1-, 2-, 5-, and 10-shots. In total, we have 40 MAPs. We run these MAPs against 100-way $\{1, 2, 3, 5, 8, 10, 16, 20,32\}$-shot tasks on those 10 datasets. For the $i^{th}$ task with a specific combination of the number of shots and target domain, we compute the relative performance ranking of these 40 MAPs: $\vec{r}_i=\{r_{ij}\},j=1,\ldots,40$ . We visualize these $\vec{r}_i$ via t-SNE in Figure~\ref{fig:tsne}, using rank-correlation $\rho$ between $\vec{r}_i$ and $\vec{r}_j$ as the distance $d=\sqrt{1-\rho(\vec{r}_i,\vec{r}_j)}$. In so doing, domains that tend to favor the same set of adaptation methods would cluster in a closer neighborhood, as shown in Figure~\ref{fig:tsne}. We have the following insightful observations.  

1) For each domain, varying the number of shots forms a continuous trajectory in the embedding space, suggesting that different adaptation methods are selected depending on the number of available labeled samples. 

2) As the number of shots increases, trajectories from multiple domains converge to several attractors (highlighted by green circles in Figure~\ref{fig:tsne}). This is consistent with the observation that finetuning gradually emerges as the most effective adaptation method for high shots.

3) DomainNet-QDraw, FGVC-Aircraft and Traffic signs datasets emerge as a distinctive cluster. One signature that sets these datasets apart from others is that they all require recognition of strokes or silhouettes as opposed to textures. This also explains why it is difficult to generalize MAPs from other DomainNet datasets to DomainNet-QDraw (see cross-domain performance in Table~\ref{table:pipeline_generalization}). Towards further improving the effectiveness of hyperparameter transfer, having a set of diverse MAPs with sufficient coverage of the embedding space is desired.


\begin{figure}
    \centering
    \includegraphics[width=1.0\textwidth]{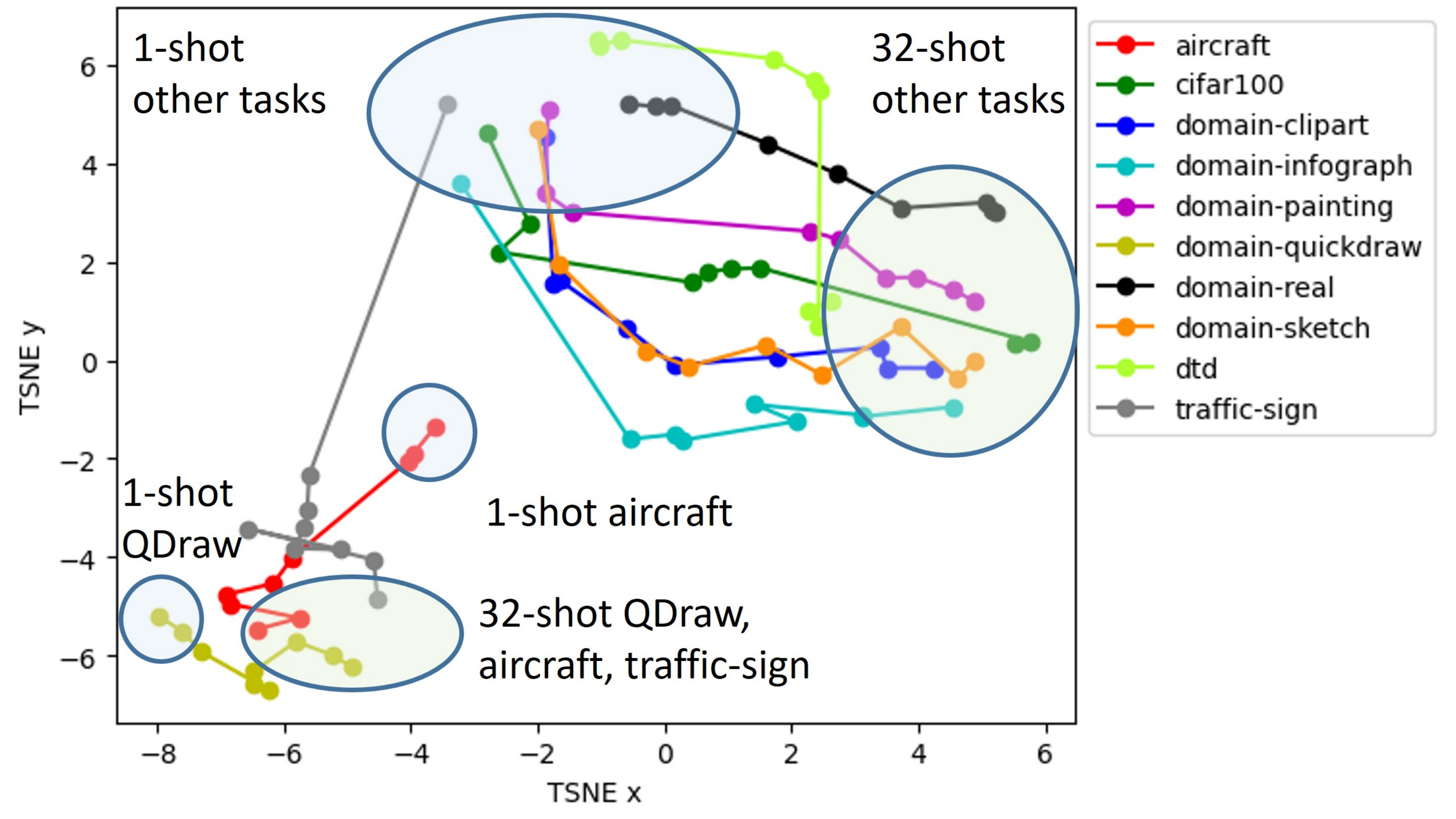}
    \label{fig:tsne}
    \caption{Domain similarity visualization using cross-domain performance of MAPs as the metric. Domains that favor similar MAPs form clusters.}
    \label{fig:tsne}
\end{figure}

\section{Conclusions}

We developed MAPs as an alternative to one-size-fits-all solutions for cross-domain FSL. Via MAP, We showed 1) that various adaptation methods can be unified under a common framework to achieve greater adaptation performance; and 2) that given a specific downstream task with limited training examples, which adaptation methods to apply can be configured and optimized dynamically and efficiently through hyperparameter transfer. Additionally, thanks to their domain specificity by design, the cross-domain performance of MAPs optimized for different domains revealed new perspectives to the similarity between domains. The effectiveness of MAPs was verified via an average $3.1\%$ improvement over finetuning on 5-shot ImageNet $\rightarrow$ 10 datasets benchmark and competitive performance against SOTA on LFT and VL3. Encouraged by these demonstrated advantages of MAP, promising directions for future research include speeding up online hyperparameter adaptation, improving hyperparameter generalization across datasets, and architectures and empirical rules for predicting hyperparameters.
\section*{Acknowledgement}

This material is based upon work supported by the United States Air Force under Contract No. FA8750-19-C-0511. Any opinions, findings and conclusions or recommendations expressed in this material are those of the author(s) and do not necessarily reflect the views of the United States Air Force.

{\small
\bibliographystyle{ieee_fullname}
\bibliography{egbib}

\begin{thebibliography}{10}\itemsep=-1pt

\bibitem{cai2020cross}
John Cai and Sheng~Mei Shen.
\newblock Cross-domain few-shot learning with meta fine-tuning.
\newblock {\em arXiv preprint arXiv:2005.10544}, 2020.

\bibitem{chen2019self}
Da Chen, Yuefeng Chen, Yuhong Li, Feng Mao, Yuan He, and Hui Xue.
\newblock Self-supervised learning for few-shot image classification.
\newblock {\em arXiv preprint arXiv:1911.06045}, 2019.

\bibitem{chen2020simple}
Ting Chen, Simon Kornblith, Mohammad Norouzi, and Geoffrey Hinton.
\newblock A simple framework for contrastive learning of visual
  representations.
\newblock In {\em ICML}, pages 1597--1607. PMLR, 2020.

\bibitem{chen2019closer}
Wei-Yu Chen, Yen-Cheng Liu, Zsolt Kira, Yu-Chiang~Frank Wang, and Jia-Bin
  Huang.
\newblock A closer look at few-shot classification.
\newblock In {\em ICLR}, 2019.

\bibitem{chen2020new}
Yinbo Chen, Xiaolong Wang, Zhuang Liu, Huijuan Xu, and Trevor Darrell.
\newblock A new meta-baseline for few-shot learning.
\newblock {\em arXiv preprint arXiv:2003.04390}, 2020.

\bibitem{cimpoi14describing}
M. Cimpoi, S. Maji, I. Kokkinos, S. Mohamed, , and A. Vedaldi.
\newblock Describing textures in the wild.
\newblock In {\em CVPR}, 2014.

\bibitem{codella2019skin}
Noel Codella, Veronica Rotemberg, Philipp Tschandl, M~Emre Celebi, Stephen
  Dusza, David Gutman, Brian Helba, Aadi Kalloo, Konstantinos Liopyris, Michael
  Marchetti, et~al.
\newblock Skin lesion analysis toward melanoma detection 2018: A challenge
  hosted by the international skin imaging collaboration (isic).
\newblock {\em arXiv preprint arXiv:1902.03368}, 2019.

\bibitem{dosovitskiy2020image}
Alexey Dosovitskiy, Lucas Beyer, Alexander Kolesnikov, Dirk Weissenborn,
  Xiaohua Zhai, Thomas Unterthiner, Mostafa Dehghani, Matthias Minderer, Georg
  Heigold, Sylvain Gelly, et~al.
\newblock An image is worth 16x16 words: Transformers for image recognition at
  scale.
\newblock {\em arXiv preprint arXiv:2010.11929}, 2020.

\bibitem{fei2006one}
Li Fei-Fei, Rob Fergus, and Pietro Perona.
\newblock One-shot learning of object categories.
\newblock {\em IEEE TPAMI}, 28(4):594--611, 2006.

\bibitem{finn2017model}
Chelsea Finn, Pieter Abbeel, and Sergey Levine.
\newblock Model-agnostic meta-learning for fast adaptation of deep networks.
\newblock In {\em International Conference on Machine Learning}, pages
  1126--1135. PMLR, 2017.

\bibitem{gidaris2019boosting}
Spyros Gidaris, Andrei Bursuc, Nikos Komodakis, Patrick P{\'e}rez, and Matthieu
  Cord.
\newblock Boosting few-shot visual learning with self-supervision.
\newblock In {\em ICCV}, pages 8059--8068, 2019.

\bibitem{grandvalet2005semi}
Yves Grandvalet, Yoshua Bengio, et~al.
\newblock Semi-supervised learning by entropy minimization.
\newblock In {\em CAP}, pages 281--296, 2005.

\bibitem{guo2020broader}
Yunhui Guo, Noel~C Codella, Leonid Karlinsky, James~V Codella, John~R Smith,
  Kate Saenko, Tajana Rosing, and Rogerio Feris.
\newblock A broader study of cross-domain few-shot learning.
\newblock In {\em ECCV}, pages 124--141. Springer, 2020.

\bibitem{guo2019spottune}
Yunhui Guo, Honghui Shi, Abhishek Kumar, Kristen Grauman, Tajana Rosing, and
  Rogerio Feris.
\newblock Spottune: transfer learning through adaptive fine-tuning.
\newblock In {\em CVPR}, 2019.

\bibitem{hariharan2017low}
Bharath Hariharan and Ross Girshick.
\newblock Low-shot visual recognition by shrinking and hallucinating features.
\newblock In {\em ICCV}, 2017.

\bibitem{helber2019eurosat}
Patrick Helber, Benjamin Bischke, Andreas Dengel, and Damian Borth.
\newblock Eurosat: A novel dataset and deep learning benchmark for land use and
  land cover classification.
\newblock {\em IEEE Journal of Selected Topics in Applied Earth Observations
  and Remote Sensing}, 12(7):2217--2226, 2019.

\bibitem{hu2020leveraging}
Yuqing Hu, Vincent Gripon, and St{\'e}phane Pateux.
\newblock Leveraging the feature distribution in transfer-based few-shot
  learning.
\newblock {\em arXiv preprint arXiv:2006.03806}, 2020.

\bibitem{kolesnikov2019big}
Alexander Kolesnikov, Lucas Beyer, Xiaohua Zhai, Joan Puigcerver, Jessica Yung,
  Sylvain Gelly, and Neil Houlsby.
\newblock Big transfer (bit): General visual representation learning.
\newblock {\em ECCV}, 2020.

\bibitem{KrauseStarkDengFei-Fei_3DRR2013}
Jonathan Krause, Michael Stark, Jia Deng, and Li Fei-Fei.
\newblock 3d object representations for fine-grained categorization.
\newblock In {\em 4th International IEEE Workshop on 3D Representation and
  Recognition (3dRR-13)}, Sydney, Australia, 2013.

\bibitem{krizhevsky2009learning}
Alex Krizhevsky, Geoffrey Hinton, et~al.
\newblock Learning multiple layers of features from tiny images.
\newblock 2009.

\bibitem{lee2013pseudo}
Dong-Hyun Lee et~al.
\newblock Pseudo-label: The simple and efficient semi-supervised learning
  method for deep neural networks.
\newblock In {\em Workshop on challenges in representation learning, ICML},
  volume~3, 2013.

\bibitem{li2020rethinking}
Hao Li, Pratik Chaudhari, Hao Yang, Michael Lam, Avinash Ravichandran, Rahul
  Bhotika, and Stefano Soatto.
\newblock Rethinking the hyperparameters for fine-tuning.
\newblock {\em ICLR}, 2020.

\bibitem{li2018explicit}
Xuhong Li, Yves Grandvalet, and Franck Davoine.
\newblock Explicit inductive bias for transfer learning with convolutional
  networks.
\newblock {\em ICML}, 2018.

\bibitem{liu2018learning}
Yanbin Liu, Juho Lee, Minseop Park, Saehoon Kim, Eunho Yang, Sung~Ju Hwang, and
  Yi Yang.
\newblock Learning to propagate labels: Transductive propagation network for
  few-shot learning.
\newblock {\em ICLR}, 2019.

\bibitem{maji13fine-grained}
S. Maji, J. Kannala, E. Rahtu, M. Blaschko, and A. Vedaldi.
\newblock Fine-grained visual classification of aircraft.
\newblock Technical report, 2013.

\bibitem{mangla2020charting}
Puneet Mangla, Nupur Kumari, Abhishek Sinha, Mayank Singh, Balaji
  Krishnamurthy, and Vineeth~N Balasubramanian.
\newblock Charting the right manifold: Manifold mixup for few-shot learning.
\newblock In {\em WACV}, pages 2218--2227, 2020.

\bibitem{mohanty2016using}
Sharada~P Mohanty, David~P Hughes, and Marcel Salath{\'e}.
\newblock Using deep learning for image-based plant disease detection.
\newblock {\em Frontiers in plant science}, 7:1419, 2016.

\bibitem{nichol2018first}
Alex Nichol, Joshua Achiam, and John Schulman.
\newblock On first-order meta-learning algorithms.
\newblock {\em arXiv preprint arXiv:1803.02999}, 2018.

\bibitem{peng2019moment}
Xingchao Peng, Qinxun Bai, Xide Xia, Zijun Huang, Kate Saenko, and Bo Wang.
\newblock Moment matching for multi-source domain adaptation.
\newblock In {\em ICCV}, pages 1406--1415, 2019.

\bibitem{perrone2018scalable}
Valerio Perrone, Rodolphe Jenatton, Matthias Seeger, and C{\'e}dric Archambeau.
\newblock Scalable hyperparameter transfer learning.
\newblock In {\em NeurIPS}, pages 6846--6856, 2018.

\bibitem{qiao2018few}
Siyuan Qiao, Chenxi Liu, Wei Shen, and Alan~L Yuille.
\newblock Few-shot image recognition by predicting parameters from activations.
\newblock In {\em CVPR}, 2018.

\bibitem{radosavovic2020designing}
Ilija Radosavovic, Raj~Prateek Kosaraju, Ross Girshick, Kaiming He, and Piotr
  Doll{\'a}r.
\newblock Designing network design spaces.
\newblock In {\em CVPR}, pages 10428--10436, 2020.

\bibitem{ravi2016optimization}
Sachin Ravi and Hugo Larochelle.
\newblock Optimization as a model for few-shot learning.
\newblock {\em ICLR}, 2017.

\bibitem{ren2018meta}
Mengye Ren, Eleni Triantafillou, Sachin Ravi, Jake Snell, Kevin Swersky,
  Joshua~B Tenenbaum, Hugo Larochelle, and Richard~S Zemel.
\newblock Meta-learning for semi-supervised few-shot classification.
\newblock {\em ICLR}, 2018.

\bibitem{satorras2018few}
Victor~Garcia Satorras and Joan~Bruna Estrach.
\newblock Few-shot learning with graph neural networks.
\newblock In {\em ICLR}, 2018.

\bibitem{snell2017prototypical}
Jake Snell, Kevin Swersky, and Richard~S Zemel.
\newblock Prototypical networks for few-shot learning.
\newblock {\em NeurIPS}, 2017.

\bibitem{sohn2020fixmatch}
Kihyuk Sohn, David Berthelot, Chun-Liang Li, Zizhao Zhang, Nicholas Carlini,
  Ekin~D Cubuk, Alex Kurakin, Han Zhang, and Colin Raffel.
\newblock Fixmatch: Simplifying semi-supervised learning with consistency and
  confidence.
\newblock {\em arXiv preprint arXiv:2001.07685}, 2020.

\bibitem{Stallkamp2012}
J. Stallkamp, M. Schlipsing, J. Salmen, and C. Igel.
\newblock Man vs. computer: Benchmarking machine learning algorithms for
  traffic sign recognition.
\newblock {\em Neural Networks}, (0):--, 2012.

\bibitem{tan2019efficientnet}
Mingxing Tan and Quoc Le.
\newblock Efficientnet: Rethinking model scaling for convolutional neural
  networks.
\newblock In {\em ICML}, pages 6105--6114. PMLR, 2019.

\bibitem{tarvainen2017mean}
Antti Tarvainen and Harri Valpola.
\newblock Mean teachers are better role models: Weight-averaged consistency
  targets improve semi-supervised deep learning results.
\newblock {\em NeuRIPS}, 2017.

\bibitem{tian2020rethinking}
Yonglong Tian, Yue Wang, Dilip Krishnan, Joshua~B Tenenbaum, and Phillip Isola.
\newblock Rethinking few-shot image classification: a good embedding is all you
  need?
\newblock {\em ECCV}, 2020.

\bibitem{touvron2020training}
Hugo Touvron, Matthieu Cord, Matthijs Douze, Francisco Massa, Alexandre
  Sablayrolles, and Herv{\'e} J{\'e}gou.
\newblock Training data-efficient image transformers \& distillation through
  attention.
\newblock {\em arXiv preprint arXiv:2012.12877}, 2020.

\bibitem{triantafillou2019meta}
Eleni Triantafillou, Tyler Zhu, Vincent Dumoulin, Pascal Lamblin, Utku Evci,
  Kelvin Xu, Ross Goroshin, Carles Gelada, Kevin Swersky, Pierre-Antoine
  Manzagol, et~al.
\newblock Meta-dataset: A dataset of datasets for learning to learn from few
  examples.
\newblock {\em ICLR}, 2020.

\bibitem{tseng2020cross}
Hung-Yu Tseng, Hsin-Ying Lee, Jia-Bin Huang, and Ming-Hsuan Yang.
\newblock Cross-domain few-shot classification via learned feature-wise
  transformation.
\newblock In {\em ICLR}, 2020.

\bibitem{van2018inaturalist}
Grant Van~Horn, Oisin Mac~Aodha, Yang Song, Yin Cui, Chen Sun, Alex Shepard,
  Hartwig Adam, Pietro Perona, and Serge Belongie.
\newblock The inaturalist species classification and detection dataset.
\newblock In {\em CVPR}, pages 8769--8778, 2018.

\bibitem{vinyals2016matching}
Oriol Vinyals, Charles Blundell, Timothy Lillicrap, Koray Kavukcuoglu, and Daan
  Wierstra.
\newblock Matching networks for one shot learning.
\newblock {\em NeurIPS}, 2016.

\bibitem{WahCUB_200_2011}
C. Wah, S. Branson, P. Welinder, P. Perona, and S. Belongie.
\newblock {The Caltech-UCSD Birds-200-2011 Dataset}.
\newblock Technical Report CNS-TR-2011-001, California Institute of Technology,
  2011.

\bibitem{wang2017chestx}
Xiaosong Wang, Yifan Peng, Le Lu, Zhiyong Lu, Mohammadhadi Bagheri, and
  Ronald~M Summers.
\newblock Chestx-ray8: Hospital-scale chest x-ray database and benchmarks on
  weakly-supervised classification and localization of common thorax diseases.
\newblock In {\em CVPR}, pages 2097--2106, 2017.

\bibitem{wang2017growing}
Yu-Xiong Wang, Deva Ramanan, and Martial Hebert.
\newblock Growing a brain: Fine-tuning by increasing model capacity.
\newblock In {\em Proceedings of the IEEE Conference on Computer Vision and
  Pattern Recognition}, pages 2471--2480, 2017.

\bibitem{Yan_2020_CVPR}
Xi Yan, David Acuna, and Sanja Fidler.
\newblock Neural data server: A large-scale search engine for transfer learning
  data.
\newblock In {\em CVPR}, June 2020.

\bibitem{ye2020hybrid}
Meng Ye, Xiao Lin, Giedrius Burachas, Ajay Divakaran, and Yi Yao.
\newblock Hybrid consistency training with prototype adaptation for few-shot
  learning.
\newblock {\em arXiv preprint arXiv:2011.10082}, 2020.

\bibitem{zamir2018taskonomy}
Amir~R Zamir, Alexander Sax, William Shen, Leonidas~J Guibas, Jitendra Malik,
  and Silvio Savarese.
\newblock Taskonomy: Disentangling task transfer learning.
\newblock In {\em Proceedings of the IEEE conference on computer vision and
  pattern recognition}, pages 3712--3722, 2018.

\bibitem{zhai2019visual}
Xiaohua Zhai, Joan Puigcerver, Alexander Kolesnikov, Pierre Ruyssen, Carlos
  Riquelme, Mario Lucic, Josip Djolonga, Andre~Susano Pinto, Maxim Neumann,
  Alexey Dosovitskiy, et~al.
\newblock The visual task adaptation benchmark.
\newblock 2019.

\bibitem{zhou2014learning}
Bolei Zhou, Agata Lapedriza, Jianxiong Xiao, Antonio Torralba, and Aude Oliva.
\newblock Learning deep features for scene recognition using places database.
\newblock In {\em NeurIPS}, pages 487--495, 2014.

\bibitem{zoph2016neural}
Barret Zoph and Quoc~V Le.
\newblock Neural architecture search with reinforcement learning.
\newblock {\em ICLR}, 2017.

\end{thebibliography}
}

\clearpage

\appendix
\setcounter{table}{0}
\renewcommand{\thetable}{S\arabic{table}}%
\setcounter{figure}{0}
\renewcommand{\thefigure}{S\arabic{figure}}%

\section{Additional Details on MAP Modules}
\label{sec:module_details}

\subsection{Finetuning} 

\paracompact{Implementation details} 
\begin{itemize}
    \item Choice of Adam/SGD+momentum optimizers.
    \item Separate learning rates for embedding and classifier layers.
    \item Choice of reinitializing the classifier layer with a new fully connected layer.
    \item Choice of data augmentation algorithms.
    \item Learning rate stepping. Decrease learning rate to $0.1\times lr$ after a portion of total runtime.
\end{itemize}

\paracompact{Hyperparameters}
\begin{itemize}
    \item switch: $\{\text{on},\text{off}\}$
    \item reinitialize: $\{\text{yes},\text{no}\}$
    \item optimizer: $\{\text{sgd},\text{adam}\}$
    \item aug: $\{\text{norm},\text{normal},\text{weak1},\text{weak2},\text{strong1},\text{strong2}\}$
    \item $\text{lr\_classifier}$: [1e-5,1e-1]
    \item $\text{lr\_embed}$: [1e-5,1e-1]
    \item step: [0.2,1.0]
    \item decay: [1e-7,1e-3]
    \item momentum: [0.7,0.99]
    \item epochs: [1,90]
    \item $\text{batch\_size}$: [8,48]
\end{itemize}

\subsection{Transductive ProtoNet (TransPN)} 

\paracompact{Implementation details} 
\begin{itemize}
    \item Add a per-dimensional power scaling layer to the embedding network, $y=\text{sgn}(x) |x|^p$ following~\cite{hu2020leveraging}.
    \item Implement prototypical networks with scaled cosine distance following~\cite{chen2020new}. This approach first computes the average embedding of all examples within that class as ``prototype'' embeddings (following ProtoNet~\cite{snell2017prototypical}). Then, the classification layer of the input network is replaced with by selecting the closest prototypes based on scaled cosine distance. 
    \item Option to perform CIPA transductive prototype embedding propagation following~\cite{ye2020hybrid} for transductive learning. Given unlabeled test examples, CIPA first predicts the class probabilities of the test examples, then use the test examples weighted by their class probabilities in the computation of the prototype embeddings to update the classifier. This process is repeated over multiple rounds using the updated classifier to refine the weights on the test examples in the prototype embeddings.
\end{itemize}

\paracompact{Hyperparameters}
\begin{itemize}
    \item switch: $\{\text{on},\text{off}\}$
    \item power scaling factor $p$: [0.2,4.0]
    \item cosine distance weight $\tau$: [5.0,32.0]
    \item CIPA switch: $\{\text{on},\text{off}\}$
    \item CIPA rounds: [1,32]
    \item CIPA weight on unlabeled examples: [0.001,10.0]
\end{itemize}

\subsection{Finetuning batchnorm layer (TuneBN)} 

\paracompact{Implementation details} 
\begin{itemize}
    \item Configure the momentum of BatchNorm layers 1) upon entry to control how much statistics are updated within the layer and 2) before output to control how much statistics are updated in all future layers.
    \item BatchNorm statistics are updated by forwarding batches of random unlabeled examples through the network for a number of iterations.  
\end{itemize}

\paracompact{Hyperparameters}
\begin{itemize}
    \item switch: $\{\text{on},\text{off}\}$
    \item momentum upon entry: [1e-5,1.0]
    \item iterations: [1,50]
    \item batch\_size: [8,48]
\end{itemize}

\subsection{Semi-supervised learning with pseudo labels (SSL-PseudoLabel)} 

\paracompact{Implementation details} 
\begin{itemize}
    \item Pseudo labels of unlabeled examples that passes a confidence threshold are weighted and used along with ground truth labels for training.
    \item Adam optimizer.
    \item Data augmentation algorithms are searched separately for labeled and unlabeled examples.
\end{itemize}

\paracompact{Hyperparameters}
\begin{itemize}
    \item switch: $\{\text{on},\text{off}\}$
    \item weight on pseudo labels: [0.0,1.0]
    \item threshold: [0.5,1.0]
    \item lr: [1e-5,0.1]
    \item epochs: [1,20]
    \item batch\_size: [8,48]
    \item aug\_labeled: $\{\text{norm},\text{normal},\text{weak1},\text{weak2},\text{strong1}$ $,\text{strong2}\}$
    \item aug\_unlabeled: $\{\text{norm},\text{normal},\text{weak1},\text{weak2},\text{strong1}$ $,\text{strong2}\}$
\end{itemize}

\subsection{Semi-supervised learning with entropy minimization (SSL-Entropy)} 

\paracompact{Implementation details} 
\begin{itemize}
    \item Entropy on unlabeled examples are used along with cross entropy on labeled examples as the objective for finetuning.
    \item Entropy on unlabeled examples are thresholded, to focus on examples with low entropy (high confidence).
    \item Adam optimizer.
\end{itemize}

\paracompact{Hyperparameters}
\begin{itemize}
    \item switch: $\{\text{on},\text{off}\}$
    \item weight on entropy: [0.0,1.0]
    \item threshold: [0,0.6]
    \item lr: [1e-5,0.1]
    \item epochs: [1,20]
    \item batch\_size: [8,48]
\end{itemize}

\subsection{Semi-supervised learning with student-teacher (SSL-MeanTeacher)} 

\paracompact{Implementation details} 
\begin{itemize}
    \item Based on SSL-PseudoLabel, but use a moving average ``teacher'' network to produce the pseudo labels.
    \item Choice of Adam/SGD+momentum optimizers.
    \item Choice of reinitializing the classifier layer with a new fully connected layer.
    \item Data augmentation algorithms are searched separately for labeled and unlabeled examples.
\end{itemize}

\paracompact{Hyperparameters}
\begin{itemize}
    \item switch: $\{\text{on},\text{off}\}$
    \item reinitialize: $\{\text{yes},\text{no}\}$
    \item optimizer: $\{\text{sgd},\text{adam}\}$
    \item weight on pseudo labels: [0.0,1.0]
    \item threshold: [0.5,1.0]
    \item lr: [1e-5,0.1]
    \item decay: [1e-7,1e-3]
    \item momentum: [0.7,0.99]
    \item epochs: [1,20]
    \item $\text{batch\_size}$: [8,48]
    \item aug\_labeled: $\{\text{norm},\text{normal},\text{weak1},\text{weak2},\text{strong1}$ $,\text{strong2}\}$
    \item aug\_unlabeled: $\{\text{norm},\text{normal},\text{weak1},\text{weak2},\text{strong1}$ $,\text{strong2}\}$
\end{itemize}

\subsection{Semi-supervised learning with FixMatch (SSL-FixMatch)} 

\paracompact{Implementation details} 
\begin{itemize}
    \item Adopt the FixMatch objective~\cite{sohn2020fixmatch} for semi-supervised finetuning. FixMatch uses pseudo labels generated using weakly augmented examples to guide predictions on strongly augmented examples.
    \item Choice of Adam/SGD+momentum optimizers.
    \item Configurable ratio of labeled/unlabeled examples for each batch.
    \item Choice of reinitializing the classifier layer with a new fully connected layer.
    \item Choice of using a moving average ``teacher'' network to produce the pseudo labels.
    \item WarmUpCosine learning rate schedule.
\end{itemize}

\paracompact{Hyperparameters}
\begin{itemize}
    \item switch: $\{\text{on},\text{off}\}$
    \item reinitialize: $\{\text{yes},\text{no}\}$
    \item optimizer: $\{\text{sgd},\text{adam}\}$
    \item weight on pseudo labels: [0.0,1.0]
    \item pseudo label threshold: [0.5,1.0]
    \item lr: [1e-5,0.1]
    \item decay: [1e-7,1e-3]
    \item momentum: [0.7,0.99]
    \item epochs: [1,20]
    \item $\text{batch\_size}$: [8,48]
    \item label/unlabeld ratio:[1:1,1:10]
\end{itemize}

\begin{table*}[htbp!]
\centering
\footnotesize
\tabcolsep=0.11cm
\noindent

\begin{tabular}{ c " l " c c c c c c | c " c c c c " c }
	\toprule
	 \multicolumn{14}{c}{100-way K-shot ImageNet $\rightarrow$ 10 datasets under \emph{from-scratch} hyperparameters }  \\
	 \midrule
	& & \multicolumn{7}{c}{DomainNet} & CIFAR & Textures & Aircraft & Signs & Overall  \\
	& Approach & QDraw & Infgrph & Sketch & Clipart & Pnting & Real & Average &   &   &   &   &    \\
\midrule
\multirow{3}{*}{2-shot} &	PN   & 22.29 &  9.24 & 24.09 & 34.99 & 33.78 & 66.48 & 31.81 & 34.66 & 41.21 & 15.48 & 30.37 & 31.26  \\
	                    &   FT   & 31.42 &  8.85 & 23.10 & 32.40 & 32.38 & 59.98 & 31.36 & 33.31 & 34.68 & 19.79 & 61.28 & 33.72  \\
                   &\textbf{MAP} & 33.90 & 11.07 & 29.19 & 41.27 & 35.32 & 66.14 & 36.15 & 35.35 & 42.36 & 14.96 & 59.30 & \textbf{36.89}  \\
\midrule
\multirow{3}{*}{5-shot} &	PN   & 31.06 & 14.27 & 33.62 & 44.82 & 47.25 & 73.57 & 40.77 & 43.27 & 51.15 & 19.11 & 40.74  & 39.89  \\
	                    &   FT   & 47.29 & 15.23 & 26.84 & 54.38 & 48.68 & 72.42 & 44.14 & 50.46 & 47.47 & 36.68 & 83.65  & 48.31  \\
                   &\textbf{MAP} & 47.33 & 17.18 & 43.03 & 57.15 & 46.72 & 74.79 & 47.70 & 43.99 & 52.04 & 40.00 & 84.19  & \textbf{50.64}  \\
		\midrule
\multirow{3}{*}{10-shot} &  PN   & 37.08 & 17.53 & 41.02 & 55.71 & 52.68 & 75.64 & 46.61 & 51.40 & 55.38 & 24.31 & 47.12 & 45.79  \\
	                    &   FT   & 57.51 & 20.45 & 47.79 & 64.47 & 58.25 & 76.35 & 54.14 & 61.11 & 52.94 & 56.85 & 97.26 & 59.30  \\
                   &\textbf{MAP} & 58.05 & 22.06 & 51.26 & 66.69 & 59.86 & 77.14 & 55.84 & 59.77 & 54.23 & 54.71 & 96.63 & \textbf{60.04}  \\
		\midrule
\multirow{3}{*}{20-shot} &  PN   & 40.50 & 21.07 & 45.65 & 60.54 & 56.63 & 77.65 & 50.34 & 54.72 & 60.15 & 28.62 & 54.28 & 49.98  \\
	           &   \textbf{FT}   & 64.54 & 26.95 & 56.54 & 73.27 & 65.04 & 80.98 & 61.22 & 67.98 & 61.96 & 73.37 & 98.84 & \textbf{66.95}  \\
                   &         MAP & 64.13 & 28.62 & 58.03 & 72.33 & 64.58 & 80.08 & 61.30 & 67.05 & 63.15 & 71.15 & 98.65 & 66.81  \\
		\bottomrule
\end{tabular}

\begin{tabular}{ c " l " c c c c c c | c " c c c c " c }
	\toprule
	 \multicolumn{14}{c}{100-way K-shot ImageNet $\rightarrow$ 10 datasets under \emph{oracle} hyperparameters }  \\
	 \midrule
	& & \multicolumn{7}{c}{DomainNet} & CIFAR & Textures & Aircraft & Signs & Overall  \\
	& Approach & QDraw & Infgrph & Sketch & Clipart & Pnting & Real & Average &   &   &   &   &    \\
\midrule
\multirow{3}{*}{2-shot} &	PN    & 22.33 &  9.70 & 26.18 & 34.06 & 35.46 & 66.05 & 32.30 & 34.86 & 44.00 & 15.50 & 30.70  & 31.88  \\
	                    &   FT    & 30.62 &  7.98 & 21.42 & 31.43 & 33.54 & 59.43 & 30.74 & 31.46 & 35.43 & 19.79 & 58.67  & 32.98  \\
                   &\textbf{MAP}  & 31.52 & 11.52 & 31.39 & 39.76 & 36.57 & 67.27 & 36.34 & 33.74 & 44.19 & 22.80 & 53.84  & \textbf{37.26}  \\
		\midrule
\multirow{3}{*}{5-shot} &  PN     & 32.30 & 14.71 & 40.85 & 33.71 & 44.87 & 73.08 & 41.06 & 44.00 & 50.85 & 20.26 & 41.35 & 40.28  \\
	                    &   FT    & 48.12 & 15.32 & 48.88 & 37.94 & 49.23 & 72.43 & 46.28 & 45.74 & 46.38 & 38.93 & 92.12 & 50.08  \\
                   &\textbf{MAP}  & 48.99 & 15.42 & 49.74 & 41.82 & 54.70 & 74.88 & 49.00 & 50.89 & 52.43 & 39.36 & 93.03 & \textbf{52.97}  \\
		\midrule
\multirow{3}{*}{10-shot} &  PN    & 37.62 & 17.71 & 44.60 & 41.22 & 52.37 & 75.16 & 46.61 & 51.21 & 55.60 & 24.04 & 47.56 & 45.81  \\
	                    &   FT    & 57.57 & 21.38 & 56.58 & 48.71 & 59.12 & 76.99 & 55.07 & 60.84 & 55.55 & 55.18 & 97.12 & 59.91  \\
                   &\textbf{MAP}  & 58.49 & 22.26 & 55.65 & 52.30 & 58.27 & 76.84 & 55.92 & 61.83 & 54.72 & 58.81 & 97.23 & \textbf{60.81}  \\
		\bottomrule
\end{tabular}

\caption{Performance of PN/FT/MAP under \emph{from-scratch} and \emph{oracle setups}. Results on DomainNet datasets are averaged to create Table 4 in the main paper. }

\label{table:full_result}
\end{table*}

\section{Per-dataset results of \emph{oracle} and \emph{from scratch} on the 10-dataset benchmark}
\label{sec:full_results}

Table~\ref{table:full_result} compares MAP with PN/FT under \emph{oracle} and \emph{from scratch} hyperparameter configurations, across $\{2,5,10,20\}$-shots on the ImageNet $\rightarrow$ 10 datasets benchmark. Table 4 in the main paper has presented the average result on DomainNet datasets in Table~\ref{table:full_result}.

\section{Setup details of LFT and VL3 benchmarks}
\label{sec:lft_vl3_details}

For MAP, PN and FT, we follow identical setups as the the 100-way ImageNet $\rightarrow$ 10 datasets experiments for 5-way 5-shot LFT and 5-way 5/20-shot VL3 experiments. 

Hyperparameters are configured using \emph{transfer} with 5-split cross validation. Due to time constraints, we reused pipelines preselected for the 100-way ImageNet $\rightarrow$ 10 datasets with Efficient-B0 backbone, but apply to 5-way tasks with ResNet10 backbones. As a result, LFT and VL3 results also demonstrate that such preselected pipelines are transferrable across backbone architecture and number of classes.

Experiments are repeated over 150 5-way K-shot splits. Mean and standard deviation are reported in Table 2 and Table 3 in the main paper. The introduction of fully automated hyperparameter configuration leads to higher computational cost. Due to limited time, we performed fewer rounds than existing literature (often 900+ rounds) and resulted in larger confidence intervals.

\end{document}